\documentclass[journal]{IEEEtran}
\pdfoutput=1
\usepackage[justification=centering]{caption}
\usepackage[cmex10]{amsmath}
\usepackage{array}
\usepackage{mdwmath}
\usepackage{mdwtab}
\usepackage{eqparbox}
\usepackage{url}
\usepackage{subfigure}
\usepackage{graphicx}
\usepackage{algorithm}  
\usepackage{multirow}
\usepackage{algpseudocode}  
\usepackage{amsmath} 
\usepackage[T1]{fontenc}
\usepackage{authblk}

\AtBeginDocument{%
	\providecommand\BibTeX{{%
			\normalfont B\kern-0.5em{\scshape i\kern-0.25em b}\kern-0.8em\TeX}}}

\def\cL{{ \mathbf{Conv}}}
\def\cP{{ \mathbf{Pooling}}}

\def\uP{{\mathbf {Up}}}
\def\uR{{\mathbf {ResGroup}}}

\newcommand{\tabincell}[2]{\begin{tabular}{@{}#1@{}}#2\end{tabular}}
\begin{document}
%
\title{Single Image Deraining via Scale-space Invariant Attention Neural Network}
%
%
%
\author{Bo Pang,
Deming Zhai,~\IEEEmembership{Member,~IEEE,}
Junjun~Jiang,~\IEEEmembership{Member,~IEEE,}
Xianming~Liu,~\IEEEmembership{Member,~IEEE}

\thanks{This work was supported by XX. (\emph{Corresponding author: Deming Zhai}).}

\IEEEcompsocitemizethanks{
\IEEEcompsocthanksitem B. Pang, D. Zhai, J. Jiang and X. Liu are with the School of Computer Science and Technology, Harbin Institute of Technology, Harbin 150001, China, and also with the Peng Cheng Laboratory, Shenzhen 518052, China  (zhaideming@hit.edu.cn; csxm@hit.edu.cn;  jiangjunjun@hit.edu.cn).}

\thanks{}}

%
%

\markboth{Journal of \LaTeX\ Class Files,~Vol.~14, No.~8, August~2015}%
{Shell \MakeLowercase{\textit{et al.}}: Bare Demo of IEEEtran.cls for IEEE Journals}

\maketitle


	\begin{abstract}
		Image enhancement from degradation of rainy artifacts plays a critical role in outdoor visual computing systems. In this paper, we tackle the notion of scale that deals with visual changes in appearance of rain steaks with respect to the camera. Specifically, we revisit multi-scale representation by scale-space theory, and propose to represent the multi-scale correlation in convolutional feature domain, which is more compact and robust than that in pixel domain. Moreover, to improve the modeling ability of the network, we do not treat the extracted multi-scale features equally, but design a novel scale-space invariant attention mechanism to help the network focus on parts of the features. In this way, we summarize the most activated presence of feature maps as the salient features. Extensive experiments results on synthetic and real rainy scenes demonstrate the superior performance of our scheme over the state-of-the-arts.
	\end{abstract}


%
\IEEEpeerreviewmaketitle

	\section{Introduction}
	
	In practical applications, tasks of outdoor scene analysis inevitably involve scenarios where images are captured under bad weather conditions, such as rainy days. Such factors cause degradation in image quality, resulting in unexpected impacts on subsequent tasks like object detection \cite{viola2001robust}, recognition \cite{tou1974pattern} and scene analysis \cite{itti1998model}. The image enhancement task that attempts to remove rain steaks is thus useful and necessary for outdoor visual system, which serves as the pre-processing step to help improve detection or recognition performance.
	
	Deraining becomes an active low-level image processing problem. Many works emerge in recent years, either video-based \cite{Li2019Video,Jiaying2018Erase}, or single-image based \cite{ren2019progressive,yang2019joint}. In this paper, we focus on the line of single image deraining, which can be formulated as an ill-posed problem. The early methods treat rain removal as a signal separation problem, relying on some prior modeling about the background layer and the rain steak layer. However, the artifact of rain steaks is inherently a kind of signal-dependent noise, features of which are intrinsically overlapped with those of the background in the feature space, making this inverse problem even harder to solve. The progress of deep learning based image restoration lights the path of single image deraining \cite{ren2019progressive,yang2019joint,Wang2019A}. The kind of data-driven approach is able to model more complicated mappings from rain images to clean ones, and thus achieves much better deraining results than the traditional model-based approach \cite{luo2015removing,li2016rain,gu2017joint}. 
	
	In real-world scenarios, rain steaks appear at different scales, depending on their distance from the camera. This leads to rain artifacts with varying sizes, background clutter and heavy occlusions, making single image deraining remain a challenging task. Some works try to handle this multi-scale effect in rain modeling. For instance, Fu \textit{et al.}  \cite{Fu2018Lightweight} construct pyramid frameworks to exploit the multi-scale knowledge for deraining. Jiang \textit{et al.}  \cite{Jiang2020Multi} propose to first generate Gaussian pyramid and then fuse the multi-scale information. These methods explicitly decompose the rain image into different pyramid levels by progressively downsampling 
	in pixel domain.  Yet, we note that this multi-scale representation approach through image pyramid is not optimal. The downsampling operation by removing large amount of pixels results in blurry artifact and resolution reduction, making the following fusion procedure not sufficiently informed about the key characteristics in identifying salient features in images.
	
	\begin{figure*}[htbp]
		\centering
		\includegraphics[width=1.0\linewidth]{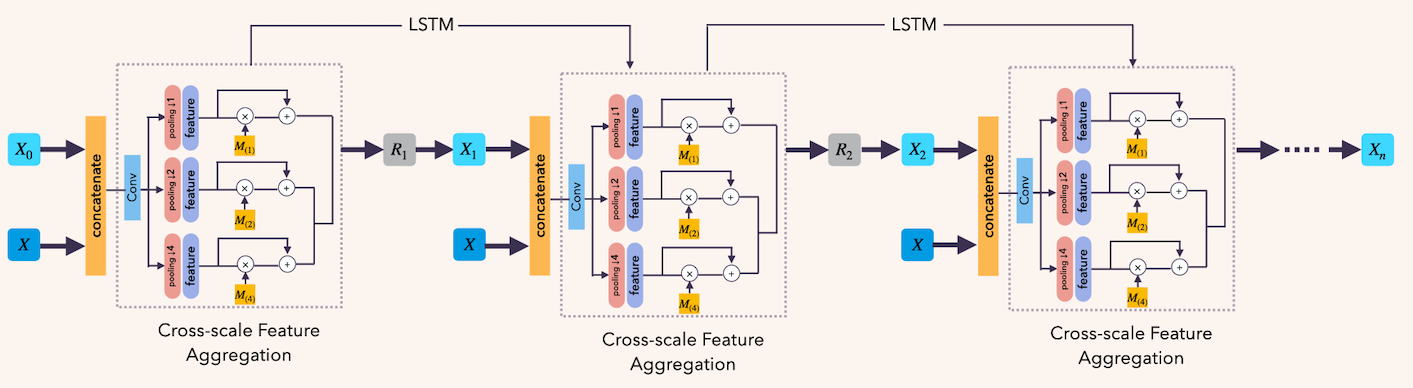}
		\caption{The overall framework of our network, which includes multiple stages. $\boldsymbol{X}$ is the input rain image, $\{\boldsymbol{X}_k\}_{k=0}^{n-1}$ are intermediate deraining results, $\boldsymbol{X}_n$ is the final deraining result, $\{\boldsymbol{R}_k\}_{k=1}^{n-1}$ are the estimated rain layers. The core module in each stage is Cross-scale Feature Aggregation, in which multi-scale features are extracted and scale-space invariant attention masks are derived. Adjacent stages are connected by LSTM to propagate information and achieve progressive refinement.}
		\label{fig:fig6}
	\end{figure*}
	
	Considering the above limitation of existing methods, in this work, we revisit multi-scale representation by scale-space theory, and propose to represent the multi-scale correlation in convolutional feature domain, which is more compact and robust than representations in pixel domain. Specifically, we propose a scale-aware deep convolutional neural network for deraining, which includes a multi-scale feature extraction branch coupling with scale-space invariant attention branch. In the feature branch, we build multi-scale pyramid in feature space through average pooling operations with various sizes, which can efficiently suppress noise and preserve background information. Besides compact representation and robustness to noise, this manner brings other benefits, such as invariance to local translation and enlarged receptive field. In the attention branch, we tailor a scale-space invariant attention mechanism to quantify the importance of the input multi-scale features to focus on. We achieve this by building difference-of-Gaussian (DoG) pyramid in scale-space, which is coupled with the feature branch and able to reveal latent salient features cross scales. Finally, LSTM-based muiti-stage refinement strategy is employed to progressively improve the network performance.
	In a nutshell, our scheme works by learning an intermediate attention maps in scale-space that are used to select the most relevant pieces of information from multi-scale features for separating the background and rain steaks.

	The main contributions of this work are highlighted as follows:
	\begin{itemize}
		\item We propose a multi-scale correlation representation in feature space for single image deraining, which is more compact and robust than the counterpart representation in pixel domain.
		\item We propose a scale-space invariant attention network through DoG pyramid, which can reveal latent salient features cross scales.
		\item Our scheme achieves the best single image deraining performance so far, which is consistently superior than a wide range of state-of-the-arts over various benchmarking datasets.
	\end{itemize}
	
	The rest sections are organized as follows: The related works are briefly overviewed in Section 2. In Section 3, we introduce the proposed method in detail. In Section 4, we provide extensive experimental results and ablation study to demonstrate the superior performance of our method.

	\section{Related Work}
	
	In this section, we briefly overview existing model-based and deep learning based single image deraining works.
	
	\subsection{Model-based methods}
	A rainy image can be modeled as a linear combination of the background layer and the rain streak layer. Based on this, model-based methods conduct deraining by explicitly defining some prior models on both the background layer and the rain streak layer. The task of deraning is then transferred to a signal separation problem. Luo \textit{et al.} \cite{luo2015removing} propose a dictionary learning method based on sparsely approximating the patches of two layers by very high discriminative codes over a learned dictionary with strong mutual exclusivity property. Li \textit{et al.} \cite{li2016rain} point out either taking dictionary learning methods or imposing a low rank structure will always leave too many rain streaks in the background image or over smooth the background image. So they propose a prior-based method using GMM which can accommodate multiple orientations and scales of the rain streaks. Gu \textit{et al.}  \cite{gu2017joint} combine analysis sparse representation (ASR) and synthesis sparse representation (SSR)  which both are used for sparsity-based image modeling proposing  a joint convolutional analysis and synthesis (JCAS) sparse representation model. However, these model-based methods couldn't well formulate complex raining process, which is not enough to retrieve background information.
	\subsection{Deep Learning based methods}
	For the data-based methods to do the rain rain removal task, the intuitive idea is to learn the mapping from the rainy image to clear backgroud image. However, such solution might cause loss of the background information. To better recover the clear image,
	Fu \textit{et al.} \cite{fu2017clearing} design a two-layers network. One is called base layer. The other is called detail layer. In base layer, it mainly focuses on low frequency information of the image using low-pass filter. While in detail layer, they design a CNN to obtain the high frequency information of the image. They enhance both the outputs of two layers and then combine them to get clean image. Due to the success of deep residual network, Fu \textit{et al.}  \cite{fu2017removing} concentrate on high frequency detail by learning the residual part between the rainy image and the clear image. Afterwards considering the different scale, direction and shape of the rain streaks,  Li \textit{et al.} \cite{li2018recurrent} adopt the dilated convolutional neural network to acquire large receptive field and used the recurrent structure. Fu \textit{et al.} \cite{Fu2018Lightweight} construct light weight pyramid frameworks to exploit the multi-scale knowledge for deraining. In Yang's work\cite{yang2019joint}, they construct contextualized dilated networks to aggregate context information at multiple scales for learning the rain features. Jiang \textit{et al.} \cite{Jiang2020Multi} propose to first generate Gaussian pyramid and then fuse the multi-scale information. In our work, we propose to represent the multi-scale correlation in convolutional feature domain, which is more compact and robust than that in pixel domain. Since there are many modules used in network, the network is too complicated to analyze each module's function. So Ren \textit{et al.} \cite{ren2019progressive} provide a simple and strong baseline using a LSTM block and several resblocks. And for better dealing with the real world's rainy image, several methods also have been proposed \cite{wang2019spatial} \cite{hu2019depth} \cite{zhang2018density}.
	
	\begin{figure}[t] 
		\centering  
		\vspace{-0.35cm} 
		\subfigtopskip=2pt 
		\subfigbottomskip=2pt 
		\subfigcapskip=-5pt 
		\subfigure[]{
			\label{level.sub.1}
			\includegraphics[width=0.32\linewidth]{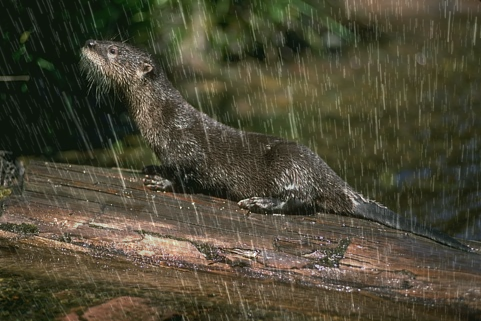}}
		\quad 
		\subfigure[]{
			\label{level.sub.2}
			\includegraphics[width=0.32\linewidth]{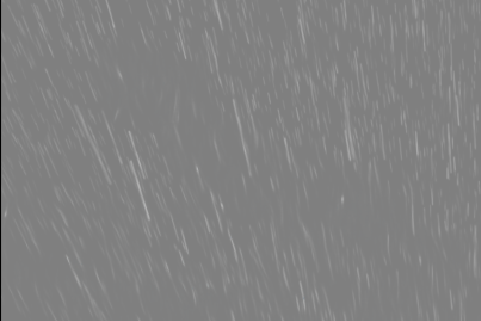}}
		
		\subfigure[]{
			\label{level.sub.3}
			\includegraphics[width=0.32\linewidth]{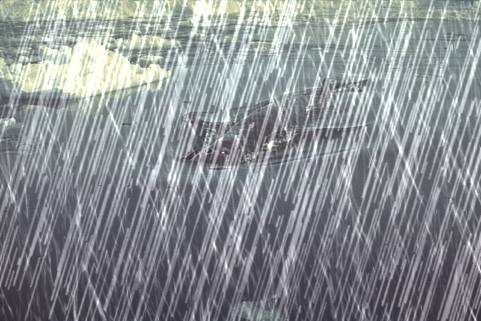}}
		\quad
		\subfigure[]{
			\label{level.sub.4}
			\includegraphics[width=0.32\linewidth]{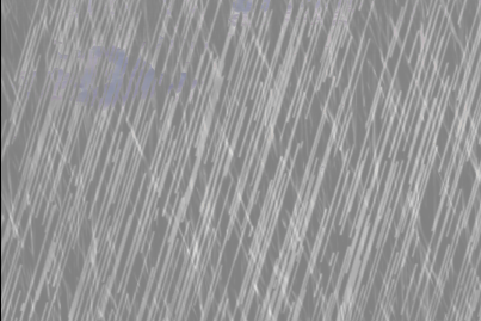}}
		\caption{Illustration of rain steak layer estimation. (a) and (c) are the input rainy images, (b) and (d) are the estimated rain streak layers by our network. }
		\label{level}
		\vspace{-0.2cm}
	\end{figure}

	\section{Proposed Method}
	
	In this section, we introduce in detail the proposed deep neural network for single image deraining. 
	
	\subsection{Network Architecture}
	The overall architecture of our proposed deraining network is illustrated in Fig. \ref{fig:fig6}. Our scheme tackles the deraining problem in a multi-stage manner, which processes the input rain image $\boldsymbol{X}$ and intermediate deraining results $\{\boldsymbol{X}_k\}_{k=0}^{n-1}$ to generate the output clean image $\boldsymbol{X}_n$ progressively, where at the beginning $\boldsymbol{X}_0 = \boldsymbol{X}$. The core module, named \textit{cross-scale feature aggregation} (CFA), is recursively conducted in each stage with the same network parameters, which is tailored to capture rich cross-scale information of input data. We connect CFA modules at adjacent stages with convolutional LSTM units that propagate information across the stages. In one CFA module, to estimate the rain steak layer $\boldsymbol{R}_k$, we learn multi-scale features $\{\boldsymbol{F}_{(s)}\}$ with three scale sizes ($\downarrow1$, $\downarrow2$, $\downarrow4$), and design a scale-space invariant attention network to derive the importance masks $\{\boldsymbol{M}_{(s)}\}$. In this way, we identify salient regions in the input data for the network to focus on, \textit{i.e.}, $\{\boldsymbol{F}^a_{(s)} = \boldsymbol{F}_{(s)}
	+ \boldsymbol{F}_{(s)}\odot \boldsymbol{M}_{(s)}\}$, which are helpful to improve the modeling ability of the proposed deraining network. The illustration of rain steak layer estimation is shown in Fig. \ref{level}. It can be found our scheme estimates the rain steak layer very well. 
	
	\subsection{Cross-scale Feature Aggregation}

	According to the scale-space theory \cite{scale_space}, the real-world objects have the nature of multi-scale, which exist as meaningful entities over certain ranges of scales. This implies that the perception of objects depends on the scale of observation. For images of unknown scenes, it is unlikely to know a prior what scales are relevant. The only reasonable approach is to represent the image data at multiple scales \cite{scale_space}. 
	
	In rain images, the notion of ``scale" deals with visual changes in rain steaks’ appearance with respect to their distance from the camera. Inspired by the scale-space theory, in the task of single image deraining, we consider to learn multi-scale features to capture a rich representation of image data. A straightforward approach is to build a coarse-to-fine pyramid in pixel domain, as done in \cite{Fu2018Lightweight} and \cite{Jiang2020Multi}. However, the image representation in pixel domain is not compact. The downsampling operation by decreasing the pixel number results in information loss about key characteristics in identifying salient features in images.
	
	\begin{figure}[t]
		\centering
		\includegraphics[width=1\linewidth]{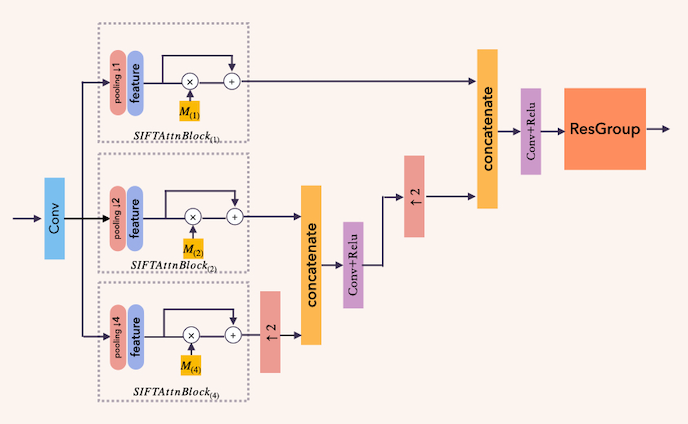}
		\caption{Cross-scale Feature Aggregation}\label{fig:fig9}
		\vspace{-0.4cm}
	\end{figure}
	
	Instead, in our work, we propose to construct multi-scale representation in feature domain, which is more compact and robust than that in pixel domain. Specifically, in the $k$-th stage, we concatenate the input rain image $\boldsymbol{X}$ and the deraining result $\boldsymbol{X}_{k-1}$ of the last stage as the input, which is fed into convolutional neural networks to extract feature maps
	\begin{equation}
	\boldsymbol{F}=\cL{(\boldsymbol{X}  \oplus \boldsymbol{X}_{k-1}) }
	\label{eq:eq1}
	\end{equation}
	where $\oplus$ denotes to the concatenation operator, $\cL{(\cdot)}$ represents one convolutional layer. Note that here we omit the parameters of $\cL{(\cdot)}$, which serve as the parameters of the overall network.
	Then we build multi-scale pyramid of CNN features through average pooling operations and one convolution layer to obtain multi-scale features:
	\begin{equation}
	\boldsymbol{F}_{(s)}=\cL{\Big( \cP{(\boldsymbol{F},s)}\Big) }
	\end{equation}
	where $\cP{(\cdot,\cdot)}$ represents the pooling operator with various size $s = 1,2,4$, which means that the receptive field of each scale is with size of $1\times1$, $2\times2$ and $4\times4$, and with a stride of 1, 2 and 4, respectively.
	
	To improve the discriminative ability of our network, we do not treat the extracted features equally, but design a novel attention mechanism that couples with the multi-scale feature extraction to help the network focus on parts of the features. In this way, we summarize the most activated presence of feature maps, which are expected to be ones of rain steaks. 
	
	As illustrated in Fig. \ref{fig:fig9}, for scale $i$, the proposed scale-space invariant attention network (SIAN) is exploited to derive the importance mask $\boldsymbol{M}_{(s)} \in [0,1]$, according to which we identify salient changes in latent CNN features:
	\begin{equation}
	\boldsymbol{F}^a_{(s)} =  \boldsymbol{F}_{(s)} + \boldsymbol{F}_{(s)} \odot {\boldsymbol{M}_{(s)} }
	\label{eq:eq2}
	\end{equation}
	where $\odot$ denotes the element-wise multiplication.
	Finally, all salient features are concatenated together to form the cross-scale feature:
	\begin{equation}
	\boldsymbol{F}^a=\boldsymbol{F}^a_{(1)} \oplus\uP\Big(\boldsymbol{F}^a_{(2)}   \oplus\uP(\boldsymbol{F}^a_{(4)} )\Big)
	\end{equation}
	where $\uP{(\cdot)}$ denotes the upsampling operator with factor 2.
	
	After one convolutional layer followed by ReLU activation, $\boldsymbol{F}^a$ is passed into $ResGroup$ which contains two Resblocks and a convolutional layer to get the estimation of rain steak layer $\boldsymbol{R}_{k}$:
	\begin{equation}
	\boldsymbol{R}_{k}=\uR(\boldsymbol{F}^a)
	\end{equation}
	The deraining result of the $k$-th stage is:
	\begin{equation}
	\boldsymbol{X_{k}}=\boldsymbol{X} -\boldsymbol{R}_{k}
	\end{equation}\par

	\subsection{Scale-space Invariant Attention Network}
	\begin{figure}[tbp]
		\centering
		\includegraphics[width=1.0\linewidth]{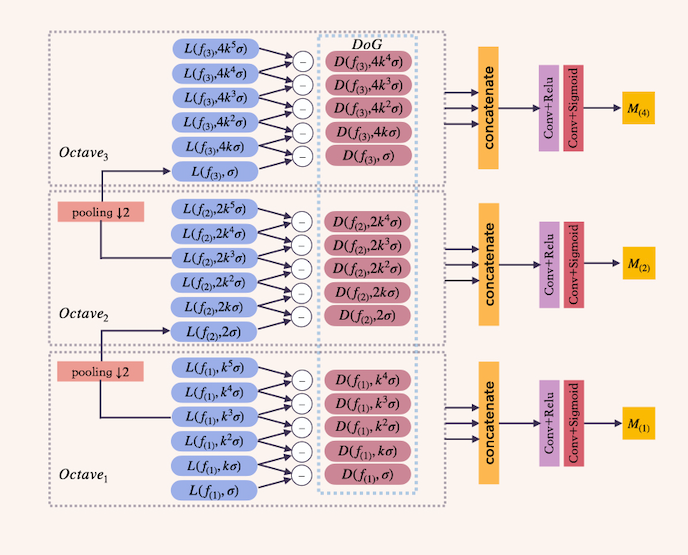}
		\caption{Scale-space Invariant Attention Network}\label{fig:fig8}
		\vspace{-0.3cm}
	\end{figure}

	In this subsection, we elaborate how we design a powerful attention mechanism that can reveal latent information from features captured at different scales. The proposed scale-space invariant attention mechanism is inspired by the classical SIFT feature extractor \cite{lowe2004sift}. SIFT achieves great success in feature extraction and description before the rise of deep learning, which is invariant to image scale and rotation, and partly invariant to affine distortion, addition of noise, and change in illumination. These wonderful properties make the underlying wisdom of SIFT particularly 
	enlightening for the task of deraining, in which the rain steaks also exhibit multi-scale, various rotations and affine transformation. The rain images also suffer from noise and low-contrast artifacts.
	
	SIFT has the four major stages of computation: 1) Scale-space extrema detection, 2) Keypoint localization, 3) Orientation assignment, 4) Keypoint descriptor. For the design of attention network, we only care about the first two stages. The keypoint, \textit{aka} extrema, which is salient change in scale-space, is naturally defined as the attention. To detect extrema, scale-space is first constructed  in CNN feature space.  As illustrated in Fig. \ref{fig:fig8}, the SIAN architecture includes three octaves, each of which corresponds to a scale-space with various smoothness levels indicating by smoothing parameter $\sigma$ and scaling parameter $sk^{l-1}$, where $l = 1,\cdots,6$ and $s = 1,2,4$. 
	
	Each octave $\boldsymbol{O}_{(s)}$ includes six layers. We define the first layer in $\boldsymbol{O}_{(s)}$ as $\boldsymbol{f}_{(s)}$. The $l$-th layer in $\boldsymbol{O}_{(s)}$ is then defined as:
	\begin{equation}
	L(\boldsymbol{f}_{(s)}, sk^{l-1}\sigma)=G(x,y,k^{l-1}\sigma) \otimes \boldsymbol{f}_{(s)}(x,y)
	\end{equation} 
	where $G(x, y,\sigma)$ is the Gaussian kernel
	\begin{equation}
	G(x, y,\sigma)=\frac{1}{2 \pi \sigma^{2}} e^{-\left(x^{2}+y^{2}\right) / 2 \sigma^{2}}
	\end{equation}
	Here $\sigma$ is set as 1.6 in practical implementation. 
	
	For the first octave $\boldsymbol{O}_{(1)}$, the first layer $\boldsymbol{f}_{(1)}$ is the Gaussian smoothed version of the feature $\boldsymbol{F}$ defined in Eq. (\ref{eq:eq1}):
	\begin{equation}
	\boldsymbol{f}_{(1)}=G(x,y,\sigma') \otimes \boldsymbol{F}
	\end{equation} 
	where $\sigma'$ is set as 1.52 in our implementation.
	For the rest two octaves, the first layer is the pooling version of the last third layer of the previous octave. Formally, the first layer of the octave $\boldsymbol{O}_{(2)}$ is 
	\begin{equation}
	\boldsymbol{f}_{(2)}=\cP{\Big(L(\boldsymbol{f}_{(1)},k^3\sigma),2\Big)}
	\end{equation} 
	and the first layer of the octave $\boldsymbol{O}_{(4)}$ is
	\begin{equation}
	\boldsymbol{f}_{(4)}=\cP{\Big(L(\boldsymbol{f}_{(2)},2k^3\sigma),2\Big)}
	\end{equation}
	Note that the pooling operation we use here is the max pooling, which is beneficial to the following local extrema detection process.
	
	\begin{figure}[t]
		\centering
		\includegraphics[width=1.0\linewidth]{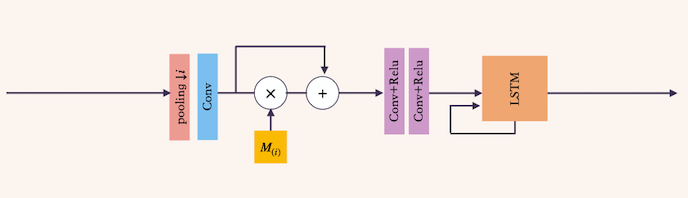}
		\caption{Progressive Refinement by LSTM}\label{fig:fig7}
	\end{figure}
	
	The derived scale-space pyramid is coupled with the multi-scale pyramid in feature extraction.  Given octaves, difference-of-Gaussian (DoG) is created by considering the difference between adjacent layers
	\begin{equation}
	D(\boldsymbol{f}_{(s)}, sk^{l}\sigma) = L(\boldsymbol{f}_{(s)}, sk^{l}\sigma) - L(\boldsymbol{f}_{(s)}, sk^{l-1}\sigma)
	\end{equation}
	According to the DoG pyramid, we then detect the local extrema, \textit{i.e.}, the salient features. We no longer do it by comparing a point against its 26 neighbors in spatial and scale domain, as done in SIFT, but turn to the learning approach. Specifically, we concatenate all the DoG layers in octave $\boldsymbol{O}_{(s)}$, which are then passed through a convolutional layer followed by a ReLU activation, and finally we use the sigmoid function to get the final attention mask $\boldsymbol{M}_{(s)}$.

	\begin{figure*}[t] 
		\centering  
		\subfigtopskip=2pt 
		\subfigbottomskip=2pt 
		\subfigcapskip=-5pt 
		\subfigure[ Rainy (21.2dB/0.727) ]{
			\label{-1}
			\includegraphics[width=0.2\linewidth]{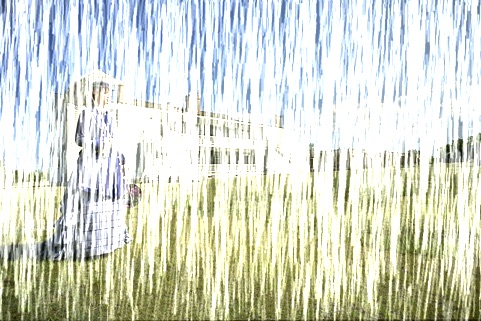}}
		\quad
		\subfigure[ Groundtruth ]{
			\label{-2}
			\includegraphics[width=0.2\linewidth]{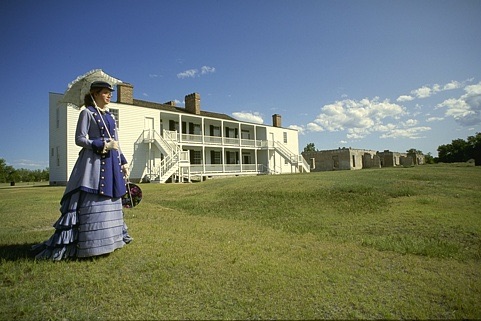}}
		\quad 
		\subfigure[ Stage=1 (19.5dB/0.704) ]{
			\label{-3l}
			\includegraphics[width=0.2\linewidth]{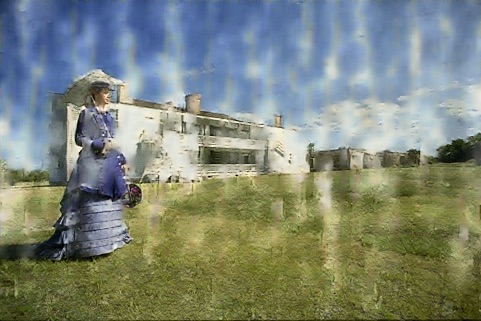}}
		\quad
		\subfigure[ Stage=2 (21.1dB/0.765)]{
			\label{-4}
			\includegraphics[width=0.2\linewidth]{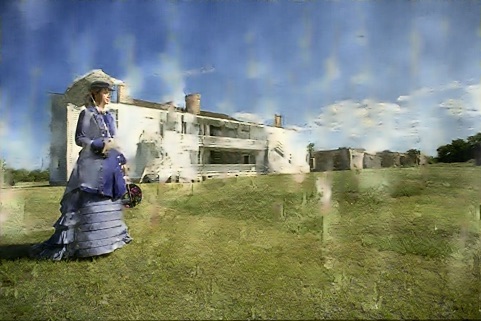}}
		
		\subfigure[ Stage=3 (22.0dB/0.787) ]{
			\label{-5}
			\includegraphics[width=0.2\linewidth]{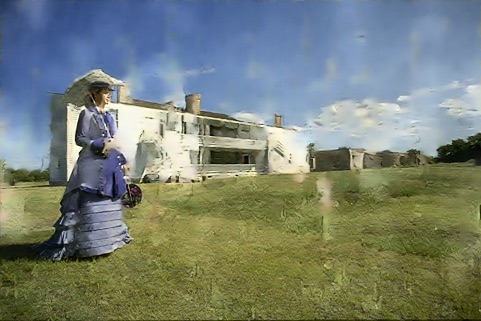}}
		\quad
		\subfigure[ Stage=4 (22.5dB/0.800)]{
			\label{-6}
			\includegraphics[width=0.2\linewidth]{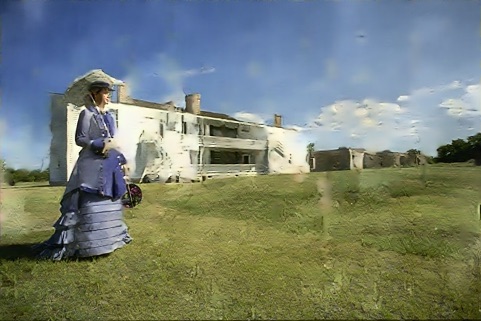}}
		\quad 
		\subfigure[ Stage=5 (22.8dB/0.804) ]{
			\label{-7}
			\includegraphics[width=0.2\linewidth]{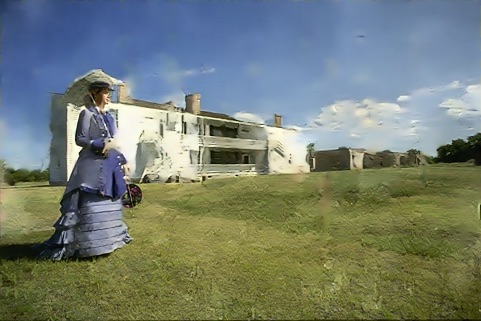}}
		\quad
		\subfigure[ Stage=6 (23.0dB/0.807) ]{
			\label{-8}
			\includegraphics[width=0.2\linewidth]{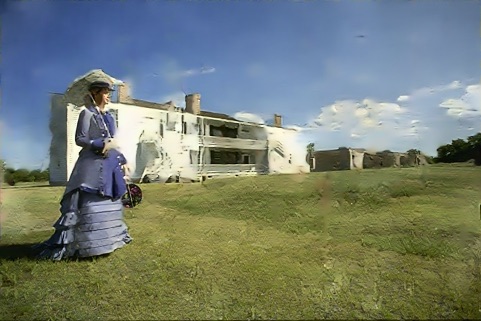}}
		
		\caption{Illustration of progressive refinement. The subjective and objective (PSNR/SSIM) results of six stages are provided. It can be found the quality of derained image is improved gradually stage-by-stage.  }
		\label{LSTM_refinement}
	\end{figure*}
	
	\begin{algorithm}[htb]  
		\caption{ Network Training Flow.}  
		\label{alg:Framwork}  
		\begin{algorithmic}[1]  
			\Require  
			Rainy images $\{X^{i}\}$;
			corresponding ground truth $\{Y^{i}\}$;
			network with initial parameter $\boldsymbol{\theta}$;
			initial learning rate $\boldsymbol{\gamma}$ ;
			\Ensure  
			Network parameters $\boldsymbol{\theta}^{*}$ 
			\For{$j$ = 1: num\_epochs} 
			\State Pick up training batch set $\{(\boldsymbol{X}^{i}$,$\boldsymbol{Y}^{i})\}^m_{i=1}$.
			\For{$i$ = 1:$m$} 
			\State $\boldsymbol{X_{0}^{i}}=\boldsymbol{X}^{i},\mathcal{L}(\boldsymbol{\theta})=0$
			\For{$k$=1:6}
			\State  $	\boldsymbol{F}=\cL{(\boldsymbol{X}^{i}  \oplus \boldsymbol{X}_{k-1}^{i}) }$;
			\State  $\boldsymbol{F_{(s)}}=\cL{\Big( \cP{(\boldsymbol{F},s)}\Big) }, s=1,2,4$;
			\State Generate attention mask $\boldsymbol{M}_{(s)}$ by SIAN;
			\State $	\boldsymbol{F}^a_{(s)} = \boldsymbol{F}_{(s)}+ \boldsymbol{F}_{(s)} \odot {\boldsymbol{M}_{(s)} }$;
			\State Update $\boldsymbol{F}^a_{(s)}$ by passing it through ConvLSTM;
			\State $ 	\boldsymbol{F}^a=\boldsymbol{F}^a_{(1)} \oplus\uP\Big(\boldsymbol{F}^a_{(2)}  \oplus\uP(\boldsymbol{F}^a_{(4)} )\Big)$;
			\State $\boldsymbol{R}_{k}^{i}=\uR(\boldsymbol{F}^a)$;
			\State $\boldsymbol{X_{k}^{i}}=\boldsymbol{X}^{i} -\boldsymbol{R}_{k}^{i}$;
			\State  $\mathcal{L}(\boldsymbol{\theta})=\mathcal{L}(\boldsymbol{\theta})-  SSIM(\boldsymbol{X_{k}^{i}},\boldsymbol{Y}^{i})$;
			\EndFor;
			\State$\boldsymbol{\theta}=\boldsymbol{\theta}-\boldsymbol{\gamma}*\mathcal{A}(\nabla_{\theta}{\mathcal{L(\boldsymbol{\theta})}})$;
			\EndFor;
			\EndFor;
			
			\State $\boldsymbol{\theta^{*}}=\boldsymbol{\theta}$.	
			
		\end{algorithmic}  
	\end{algorithm}

	\subsection{Progressive Refinement by LSTM }

	To further improve the network performance, similar to PReNet \cite{ren2019progressive}, we connect adjacent stages with convolutional LSTM (ConvLSTM) units that propagate information from the previous stage \cite{xingjian2015convolutional}. LSTM \cite{hochreiter1997long} is good at handling time-sequence data. There are three gates in LSTM, including the input gate, the forget gate, the output gate and the cell state. The key in ConvLSTM is the cell state, which encodes the state information that will be propagated to the next LSTM. In our work, 
	as illustrated in Fig. \ref{fig:fig7}, after obtaining the $\boldsymbol{F}^a_{(s)}$ according to Eq. (\ref{eq:eq2}), it is updated by passing two convolutional layers, and serves as the input of the next ConvLSTM block. In Fig. \ref{LSTM_refinement}, we provide an illustration of the effect of progressive refinement by LSTM. It can found the quality of derained image becomes better and better along with the stage number increases.

	\begin{table}[t]
		\caption{Benchmarking datasets used for network training as well as performance evaluation }
		\vspace{-0.3cm}
		\label{dataset}
		\begin{center}
			\setlength{\tabcolsep}{0.8mm}{
				\small
				\begin{tabular}{|c|c|c|}
					\hline Datasets & Sample Number (Train/Test) & Description\\\hline
					\hline Rain12 \cite{li2016rain} & 12 & \small{Only for testing} \\
					\hline Rain100L \cite{yang2017deep} & 200/100 & \tabincell{c}{Synthesized with \\ one type of  rain streaks \\ (light rain case)} \\
					\hline Rain100H \cite{yang2017deep} & 1,800/100 & \tabincell{c}{Synthesized with \\ ﬁve types of  rain streaks \\ (heavy rain case)} \\
					\hline Rain1400 \cite{fu2017removing} & 12,600/1,400 & \tabincell{c}{1,000 clean image \\ used to synthesize \\ 14,000 rain images.}\\\hline
				\end{tabular}
			}
		\end{center}
		\vspace{-0.3cm}
	\end{table}

	\begin{table}[htbp]
		\caption{Performance comparison of different loss functions}
		\vspace{-0.3cm}
		\begin{center}
			\setlength{\tabcolsep}{3mm}{
				\begin{tabular}{|c|c|c|c|c|}
					\hline
					\multirow{2}{*}{\textbf{Loss}} & \multicolumn{2}{c|}{\textbf{Rain100L}}                                  & \multicolumn{2}{c|}{\textbf{Rain100H}}                                 \\ \cline{2-5} 
					& \multicolumn{1}{c|}{\textbf{PSNR}} & \multicolumn{1}{c|}{\textbf{SSIM}} & \multicolumn{1}{c|}{\textbf{PSNR}} & \multicolumn{1}{c|}{\textbf{SSIM}} \\ \hline
					MAE                               & 38.54                              & 0.981                              & 30.12                              & 0.904                             \\ \hline
					MSE                              & 38.52                              & 0.981                              & 30.03                              & 0.902                             \\ \hline
					Negative SSIM                    & \textbf{38.80}                     & \textbf{0.984}                     & \textbf{30.33}                     & \textbf{0.909}                    \\ \hline
				\end{tabular}
			}
		\end{center}
		\label{loss}
		\vspace{-0.3cm}
	\end{table}
	
	\vspace{-0.4cm}
	\subsection{Network Training}
	The network parameters $\boldsymbol{\theta}$  involve the kernel weights of the performed convolutional layers. The training process is conducted based on several public benchmarking dataset, as shown in Table \ref{dataset}, which are synthesized data and thus there are many pairs of rain images $\{\boldsymbol{X}^i\}$ and the corresponding ground truth $\{\boldsymbol{Y}^i\}$.

	For each image, we compute its accumulated negative SSIM loss \cite{Wang2004Image} over all outputs of $K$ stages. As shown in Table \ref{loss}, negative SSIM loss works better than the popular MAE and MSE loss. Traversing all training samples, the final training loss is:
	\begin{equation}
	\mathcal{L}(\boldsymbol{\theta})=-\sum_{i=1}^{N}\sum_{k=1}^{K} SSIM(\boldsymbol{X}^i_k,\boldsymbol{Y}^i)
	\end{equation}
	where $N$ is the number of samples used for training.
	
	The optimal parameters $\boldsymbol{\theta}^{*}$ can be obtained by:
	\begin{equation}
	\boldsymbol{\theta}^{*} =\arg \min _{\boldsymbol{\theta}} \mathcal{L(\boldsymbol{\theta})}
	\end{equation}
	This minimization problem can be addressed by ADAM \cite{Kingma2014Adam}:
	\begin{equation}
	\boldsymbol{\theta}=\boldsymbol{\theta}-\boldsymbol{\gamma}*\mathcal{A}(\nabla_{\theta}{\mathcal{L(\boldsymbol{\theta})}})
	\end{equation}
	where $\boldsymbol{\gamma}$ is the learning rate, and $\mathcal{A}(\nabla_{\theta}{\mathcal{L(\boldsymbol{\theta})}})$ represents the updated value based on ADAM. The whole network training flow is summarized in Algorithm \ref{alg:Framwork}.
	
	Our network is trained on NVIDIA GTX 1080Ti. The image patch size of all the dataset are set as $120\times120$ and the batch size are set as 18. We extract image patches with stride 40, 80, and 100 for Rain100L, Rain100H and Rain1400, respectively. For Rain100L, we perform the data augmentation by horizontal flip. The training epoch of Rain100L, Rain100H and Rain1400 are set as 100, 100 and 50, respectively.  The learning rate $\boldsymbol{\gamma}$ is set as 2e-4.
	
	\section{Experiments}
	In this section, extensive quantitative and qualitative results are provided to demonstrate the superior performance of the proposed method. Ablation study is also offered to promote deeper understanding of our network.
	
	\subsection{Evaluation on Synthetic Datasets}
	
	
	\subsubsection{Comparison with the state-of-the-arts} 
	Our method is comprehensively compared with state-of-the-art model-based and deep learning based works on synthetic benchmarking datasets shown in Table \ref{dataset}. The comparison study group includes:
		\begin{figure*}[t] 
		\centering  
		\vspace{-0.35cm} 
		\subfigtopskip=2pt 
		\subfigbottomskip=2pt 
		\subfigcapskip=-5pt 
		\subfigure[ Input (21.17dB/0.727)]{
			\label{rain_100l}
			\includegraphics[width=0.18\linewidth]{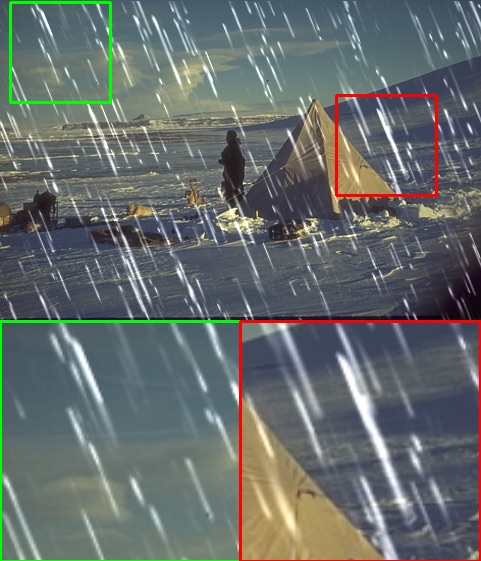}}
		\quad
		\subfigure[JCAS (24.33dB/0.809)]{
			\label{jcas_100l}
			\includegraphics[width=0.18\linewidth]{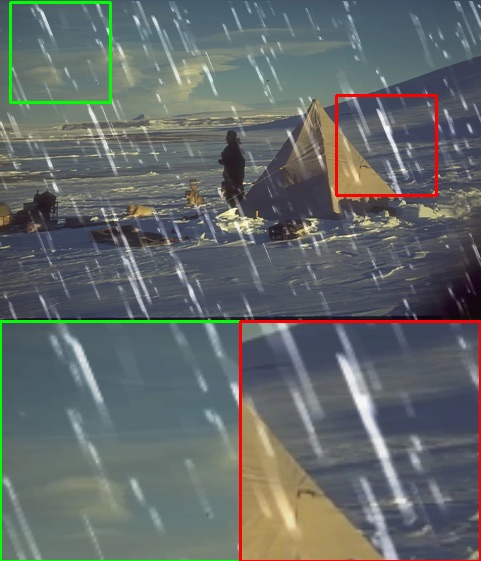}}
		\quad 
		\subfigure[LPNet (24.60dB/0.876)]{
			\label{lpnet_100l}
			\includegraphics[width=0.18\linewidth]{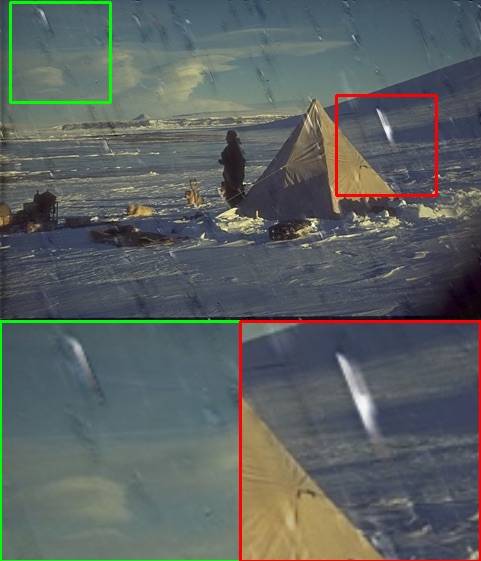}}
		\quad
		\subfigure[\scriptsize JORDER\_E(42.49dB/0.988)]{
			\label{jorder_100l}
			\includegraphics[width=0.18\linewidth]{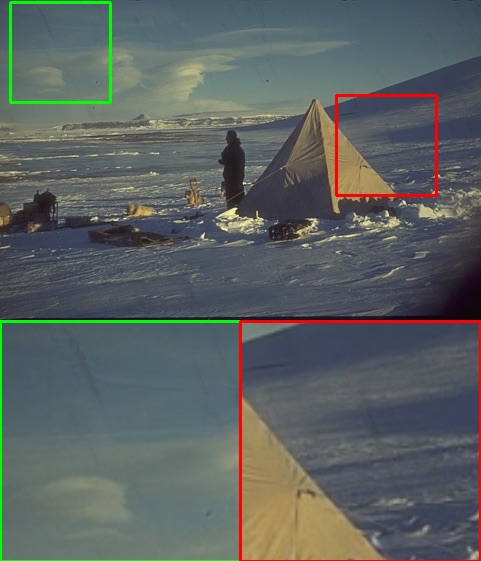}}
		
		\subfigure[PreNet (40.23dB/0.987)]{
			\label{prenet_100l}
			\includegraphics[width=0.18\linewidth]{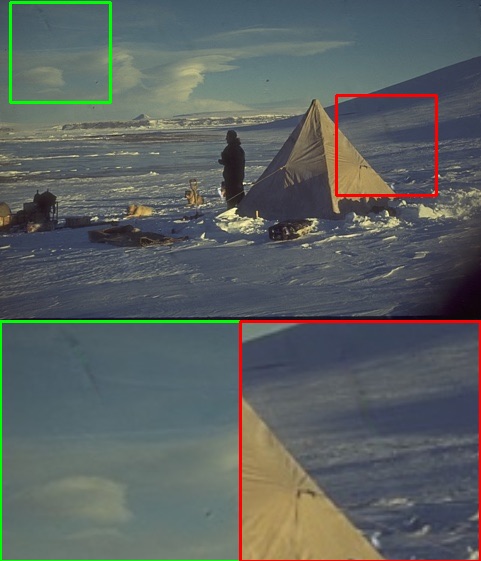}}
		\quad 
		\subfigure[RESCAN(40.98dB/0.987)]{
			\label{rescan_100l}
			\includegraphics[width=0.18\linewidth]{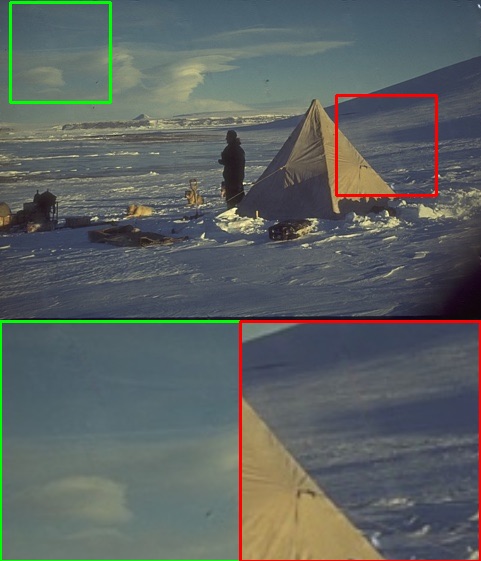}}
		\quad
		\subfigure[Ours (42.93dB/0.991)]{
			\label{sift_100l}
			\includegraphics[width=0.18\linewidth]{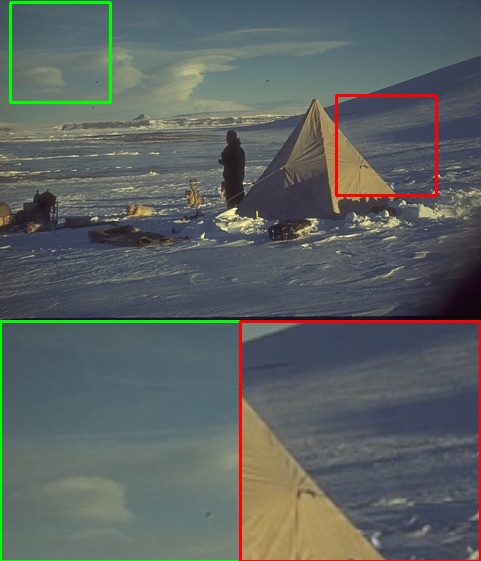}}
		\quad
		\subfigure[Groundtruth]{
			\label{gt}
			\includegraphics[width=0.18\linewidth]{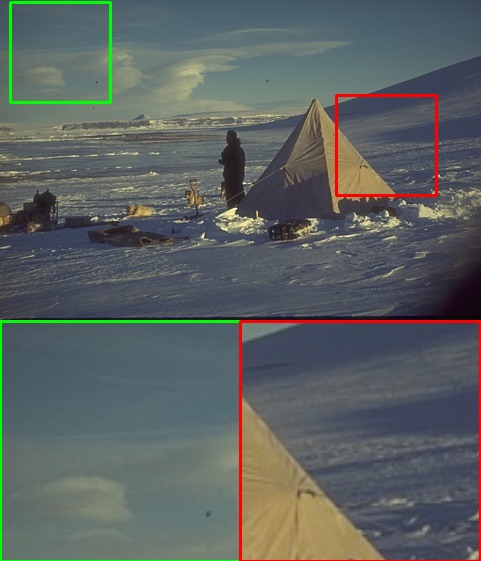}}
		\caption{Visual deraining results on sample from  Rain100L.}
		\label{Rain100L}
	\end{figure*}

	\begin{figure*}[!h] 
		\centering  
		\vspace{-0.35cm} 
		\subfigtopskip=2pt 
		\subfigbottomskip=2pt 
		\subfigcapskip=-5pt 
		\subfigure[Input (13.55dB/0.482)]{
			\label{rain_100h}
			\includegraphics[width=0.18\linewidth]{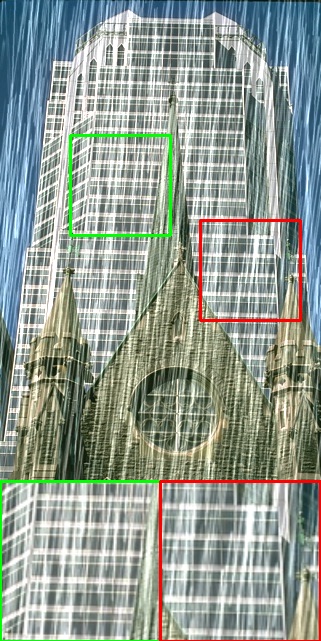}}
		\quad
		\subfigure[JCAS (15.62dB/0.570) ]{
			\label{jcas_100h}
			\includegraphics[width=0.18\linewidth]{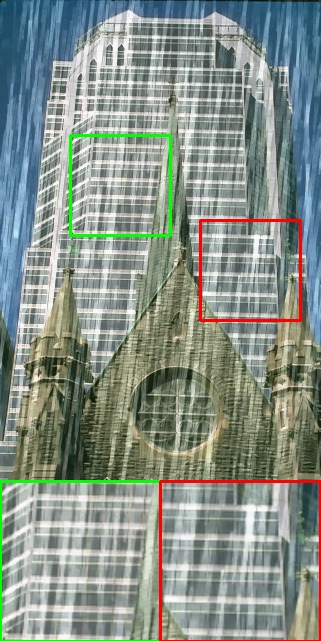}}
		\quad 
		\subfigure[LPNet (28.03dB/0.916)]{
			\label{gmm_100h}
			\includegraphics[width=0.18\linewidth]{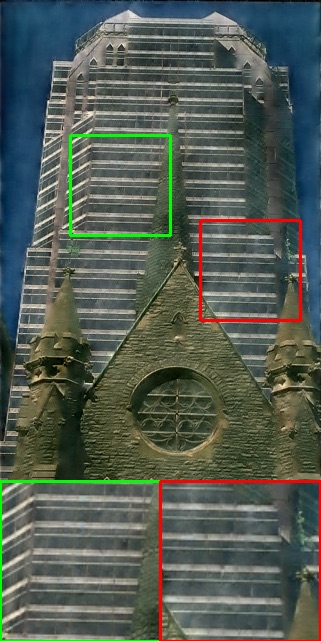}}
		\quad
		\subfigure[\scriptsize JORDER\_E (24.95dB/0.878)]{
			\label{jorder_100h}
			\includegraphics[width=0.18\linewidth]{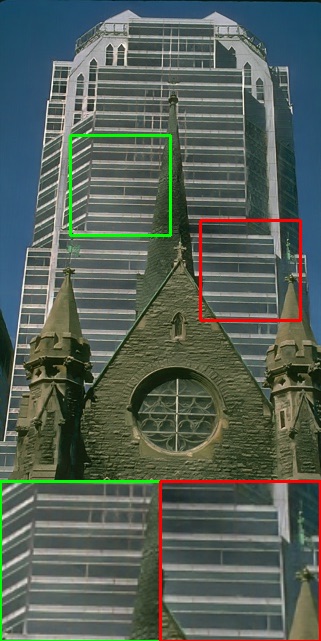}}
		
		\subfigure[PreNet (24.39dB/0.863)]{
			\label{prenet_100h}
			\includegraphics[width=0.18\linewidth]{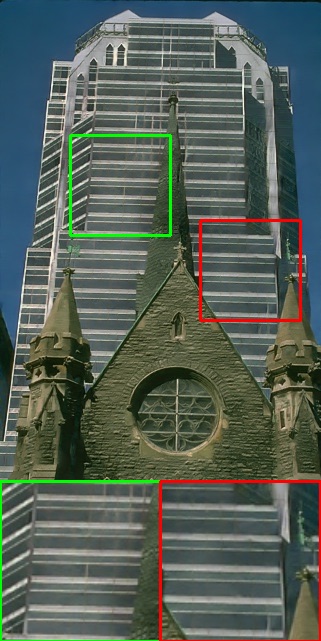}}
		\quad 
		\subfigure[RESCAN(23.31dB/0.862)]{
			\label{rescan_100h}
			\includegraphics[width=0.18\linewidth]{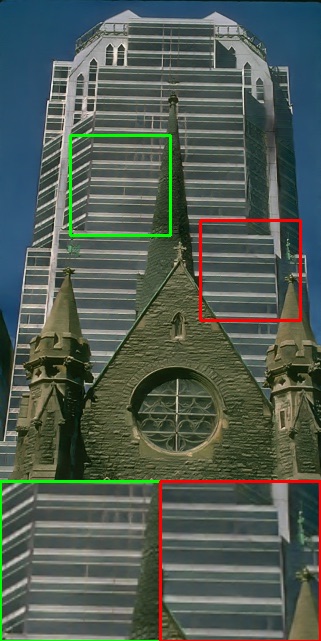}}
		\quad
		\subfigure[Ours (28.03dB/0.916)]{
			\label{sift_100h}
			\includegraphics[width=0.18\linewidth]{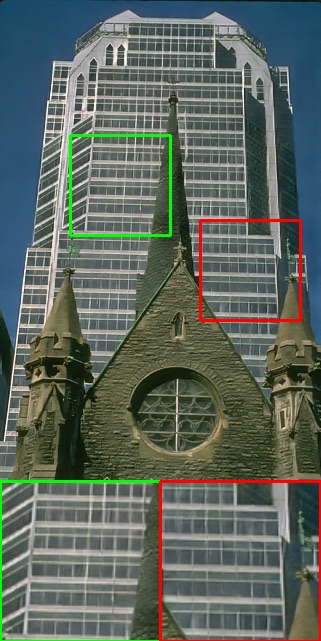}}
		\quad
		\subfigure[Groundtruth]{
			\label{gt_100h}
			\includegraphics[width=0.18\linewidth]{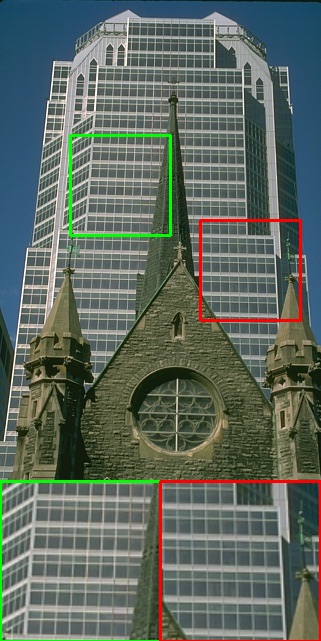}}
		
		\caption{Visual deraining results on sample from  Rain100H.}
		\label{Rain100H}
	\end{figure*}
	\begin{itemize}
		\item Model-based: 1) Discriminative Sparse Coding, DSC \cite{luo2015removing}; 2) Gaussian Mixture Model, GMM \cite{li2016rain}; 3) Joint Convolutional Analysis and Synthesis Sparse Representation, JCAS \cite{gu2017joint};
		\item Deep Learning based: 1) Clear \cite{fu2017clearing};  2) Deep Detail Network, DDN \cite{fu2017removing}; 3) Recurrent Squeeze-and-Excitation Context Aggregation Net, RESCAN \cite{li2018recurrent}; 4) Progressive Recurrent Network, PReNet \cite{ren2019progressive}; 5) Spatial Attentive Network, SPANet \cite{wang2019spatial}; 6) Enhanced JOint Rain DEtection and Removal, JORDER-E \cite{yang2019joint}; 7) Semi-supervised Image Rain Removal, SSIR \cite{wei2019semi}; 8) Lightweight Pyramid Networks, LPNet \cite{Fu2018Lightweight}.
	\end{itemize}
	
	
	We follow the same experiment settings as introduced in \cite{Wang2019A} \cite{yang2019single}. Peak signal-to-noise ratio (PSNR) and SSIM are used for quantitative performance evaluation. We only consider the luminance channel, since it has the most significant impact on the human visual system to evaluate the image quality. We adopt the numerical results reported in \cite{Wang2019A}.
	
	The quantitative evaluation results on Rain100L, Rain100H, and Rain1400, Rain12 are presented in Table \ref{100LA100H} and Table \ref{1400A12}, respectively. It can be found that, compared with deep learning based methods, three model-based methods---DSC, GMM and JCAS---achieve relatively lower PSNR and SSIM values, due to the lack of modeling ability. Deep learning based methods achieve great success in single image deraining. For instance, compared with the best performed model-based method GMM, the method JORDER\_E improves the PSNR by over 9dB on Rain100L. Among all compared data-driven methods, our proposed scheme achieves the best PSNR and SSIM performance on various datasets. The PSNR gains over the second best performed work are 0.19dB, 0.22dB, 0.12dB and 0.55dB on Rain100L, Rain100H, Rain1400 and Rain12, respectively. These results demonstrate the superiority of our work. 
	
\begin{figure*}[t] 
		\centering  
		\vspace{-0.35cm} 
		\subfigtopskip=2pt 
		\subfigbottomskip=2pt 
		\subfigcapskip=-5pt 
		\subfigure[Input (21.59dB/0.771)]{
			\label{rain_1400}
			\includegraphics[width=0.20\linewidth]{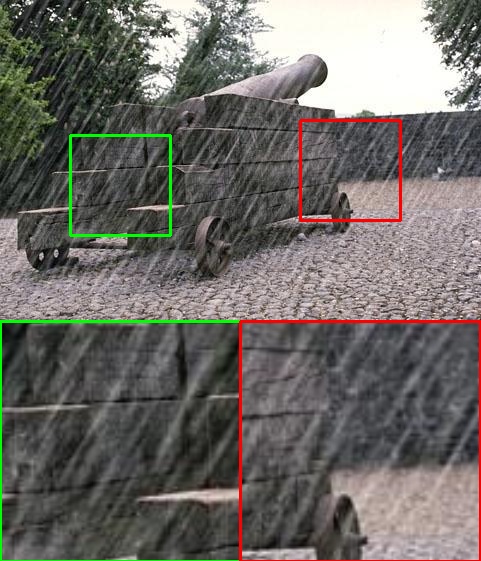}}
		\quad 
		\subfigure[PreNet (29.76dB/0.908)]{
			\label{prenet_1400}
			\includegraphics[width=0.20\linewidth]{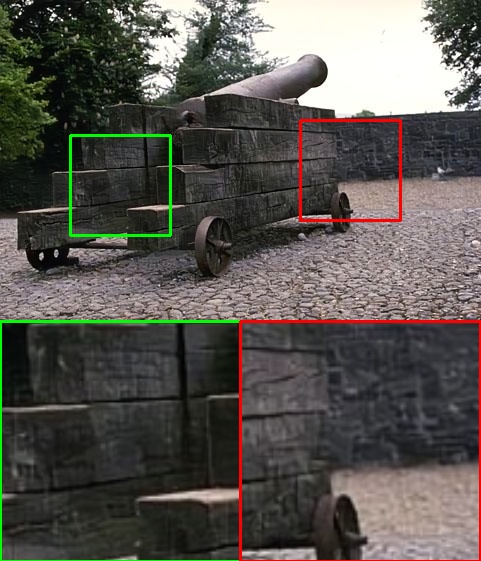}}
		\quad
		\subfigure[Ours (29.93dB/0.908)]{
			\label{sift_1400}
			\includegraphics[width=0.20\linewidth]{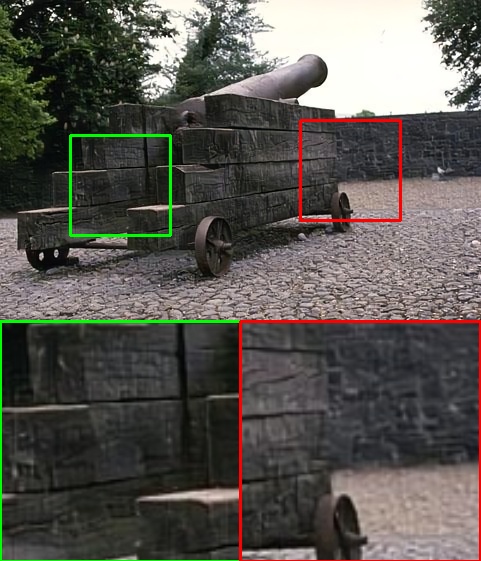}}
		\quad
		\subfigure[Groundtruth]{
			\label{gt_1400}
			\includegraphics[width=0.20\linewidth]{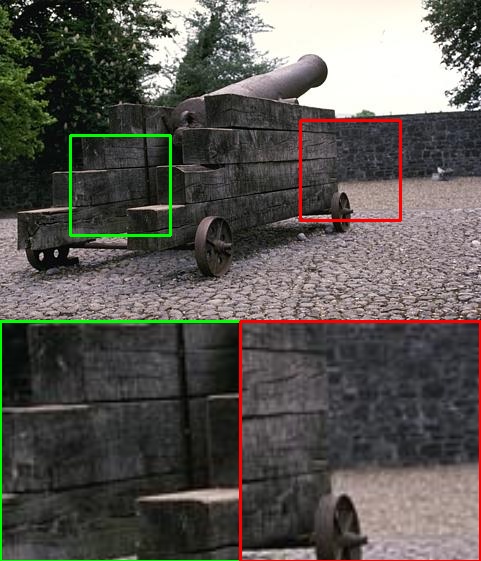}}
		
		\subfigure[Input (28.48dB/0.798)]{
			\label{rain_12}
			\includegraphics[width=0.20\linewidth]{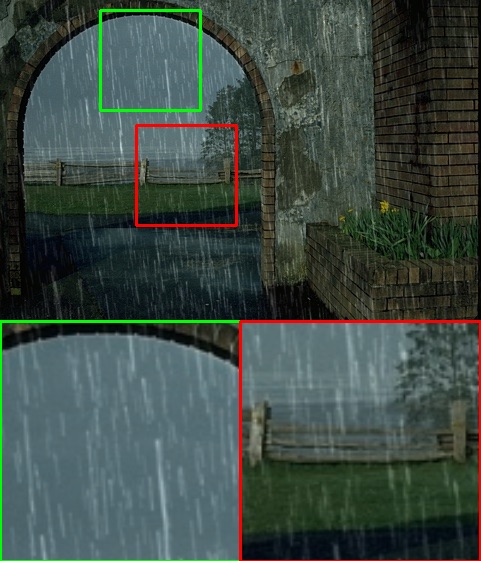}}
		\quad 
		\subfigure[PreNet (35.22dB/0.936)]{
			\label{prenet_12}
			\includegraphics[width=0.20\linewidth]{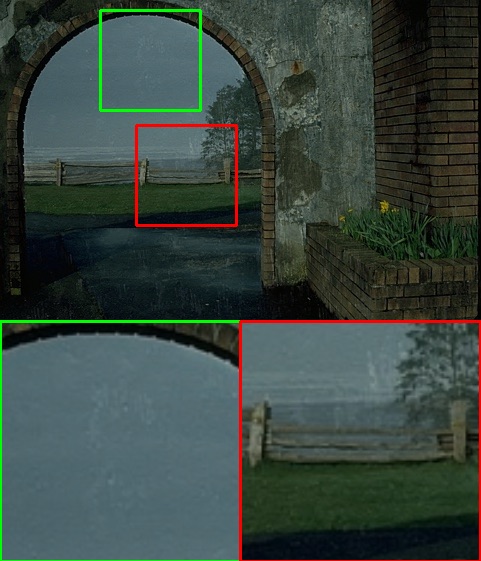}}
		\quad
		\subfigure[Ours (36.08dB/0.948)]{
			\label{sift_12}
			\includegraphics[width=0.20\linewidth]{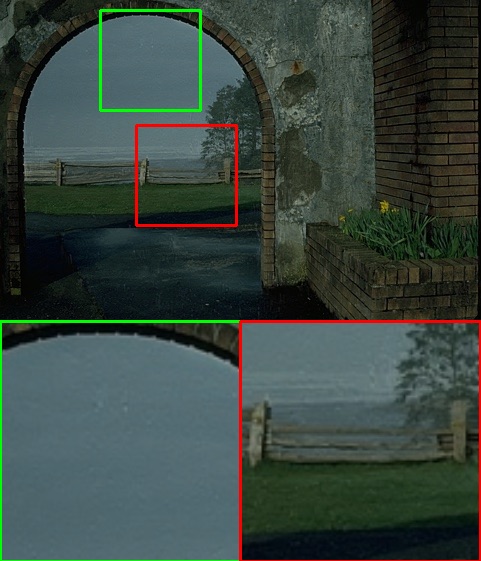}}
		\quad
		\subfigure[Groundtruth]{
			\label{gt_12}
			\includegraphics[width=0.20\linewidth]{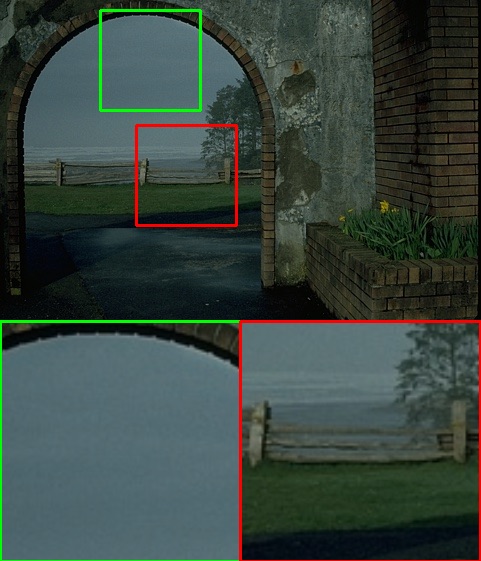}}
		
		\caption{Visual deraining results on samples from Rain1400 and Rain12.}
		\label{Rain1400A12}
	\end{figure*}

	We also provide qualitative evaluation through visual quality comparison. The example images cover various scenarios, including light rain steaks, large rain streaks and dense rain accumulation. As illustrated in Fig. \ref{Rain100L}-Fig. \ref{Rain1400A12}, the model-based method JCAS cannot remove the rain steaks well. In its deraining results, most rain steaks are still existing. LPNet, which builds coarse-to-fine pyramid in pixel domain to exploit the multi-scale correlation, cannot preserve the image structures well. It can be found in Fig. \ref{Rain100L}, LPNet cannot remove heavy rain steaks. PreNet also employs the progressive refinement manner as ours. However, as shown in Fig. \ref{Rain100H}, it leads to oversmoothing effect in the building regions. JORDER\_E performs the second best in quantitative evaluation. In subjective evaluation, it can be seen there is still rain trace in sky region of Fig. \ref{Rain100L}; it also suffers from oversmoothing as PReNet in Fig. \ref{Rain100H}.
	
	\begin{table}[t]
		\caption{The quantitative evaluation results with respect to PSNR (dB)/SSIM on Rain100L and Rain100H. The best and the second ones are highlighted by bold and underline. }
		\vspace{-0.3cm}
		\begin{center}
			\setlength{\tabcolsep}{2mm}{
				\begin{tabular}{c|cc|cc}
					\hline Datasets & \multicolumn{2}{|c|} { Rain100L } & \multicolumn{2}{c} { Rain $100 \mathrm{H}$} \\
					\hline Metrics & PSNR & SSIM & PSNR & SSIM \\
					\hline Input & 26.90 & 0.838 & 13.56 & 0.371 \\
					\hline DSC \cite{luo2015removing} (ICCV'15) & 27.34 & 0.849 & 13.77 & 0.320 \\
					\hline GMM \cite{li2016rain} (CVPR'16) & 29.05 & 0.872 & 15.23 & 0.450 \\
					\hline JCAS \cite{gu2017joint} (ICCV'17) & 28.54 & 0.852 & 14.62 & 0.451 \\
					\hline Clear \cite{fu2017clearing} (TIP'17) & 30.24 & 0.934 & 15.33 & 0.742 \\
					\hline DDN \cite{fu2017removing}(CVPR'17) & 32.38 & 0.926 & 22.85 & 0.725 \\
					\hline RESCAN \cite{li2018recurrent} (ECCV'18) & 38.52 & 0.981 & 29.62 & 0.872 \\
					\hline PReNet \cite{ren2019progressive} (CVPR'19) & 37.45 & 0.979 & $\underline{30.11}$ & 0.905 \\
					\hline SPANet \cite{wang2019spatial} (CVPR'19) & 34.46 & 0.962 & 25.11 & 0.833 \\
					\hline JORDER\_E \cite{yang2019joint} (TPAMI'19) & $\underline{3 8.6 1}$ & $\underline{0 . 9 8 2}$ & 30.04 & $\underline{0 . 9 0 6}$ \\
					\hline SIRR \cite{wei2019semi} (CVPR'19) & 32.37 & 0.926 & 22.47 & 0.716 \\
					\hline LPNet \cite{Fu2018Lightweight} (TNNLS' 20) & 33.40 & 0.960  & 23.40 & 0.820 \\
					\hline Ours  & \textbf{38.80} & \textbf{0.984} & \textbf{30.33} & \textbf{0.909} \\
					\hline
				\end{tabular}
			}
		\end{center}
		\vspace{-0.3cm}
		\label{100LA100H}
	\end{table}

	\begin{table}[t]
		\caption{The quantitative evaluation results with respect to PSNR (dB)/SSIM on Rain1400 and Rain12. The best and the second ones are highlighted by bold and underline.}
		\vspace{-0.3cm}
		\begin{center}

			\setlength{\tabcolsep}{2mm}{
				\begin{tabular}{c|cc|cc}
					\hline Datasets & \multicolumn{2}{|c|} { Rain1400 } & \multicolumn{2}{c} { Rain12 } \\
					\hline Metrics & PSNR & SSIM & PSNR & SSIM \\
					\hline Input & 25.24 & 0.810 & 30.14 & 0.856 \\
					\hline DSC \cite{luo2015removing}  (ICCV'15) & 27.88 & 0.839 & 30.07 & 0.866 \\
					\hline GMM  \cite{li2016rain}  (CVPR'16) & 27.78 & 0.859 & 32.14 & 0.916 \\
					\hline JCAS \cite{gu2017joint} (ICCV'17) & 26.20 & 0.847 & 33.10 & 0.931 \\
					\hline Clear \cite{fu2017clearing}  (TIP'17) & 26.21 & 0.895 & 31.24 & 0.935 \\
					\hline DDN \cite{fu2017removing} (CVPR'17) & 28.45 & 0.889 & 34.04 & 0.933 \\
					\hline RESCAN  \cite{li2018recurrent} (ECCV'18) & 32.03 & 0.931 & 36.43 & 0.952 \\
					\hline PReNet \cite{ren2019progressive} (CVPR'19) & 32.55 & $\underline{0 . 9 4 6}$ & 36.66 & 0.961 \\
					\hline SPANet \cite{wang2019spatial} (CVPR'19) & 29.76 & 0.908 & 34.63 & 0.943 \\
					\hline JORDER\_E \cite{yang2019joint} (TPAMI'19) & $\underline{3 2 . 6 8}$ & 0.943 & $\underline{3 6 . 6 9}$ & $\underline{0 . 9 6 2}$ \\
					\hline SIRR \cite{wei2019semi} (CVPR'19) & 28.44 & 0.889 & 34.02 & 0.935 \\
					\hline LPNet \cite{Fu2018Lightweight} (TNNLS' 20) & - & -  & 34.7 & 0.95 \\
					\hline Ours & \textbf{32.80} & \textbf{0.946} & \textbf{37.24} & \textbf{0.967} \\
					\hline
				\end{tabular}
			}
		\end{center}
		\vspace{-0.3cm}
		\label{1400A12}
	\end{table}

	\begin{figure*}[t] 
		\centering  
		\vspace{-0.35cm} 
		\subfigtopskip=2pt 
		\subfigbottomskip=2pt 
		\subfigcapskip=-5pt 
		\subfigure[ Input ]{
			\label{rain_real}
			\includegraphics[width=0.20\linewidth]{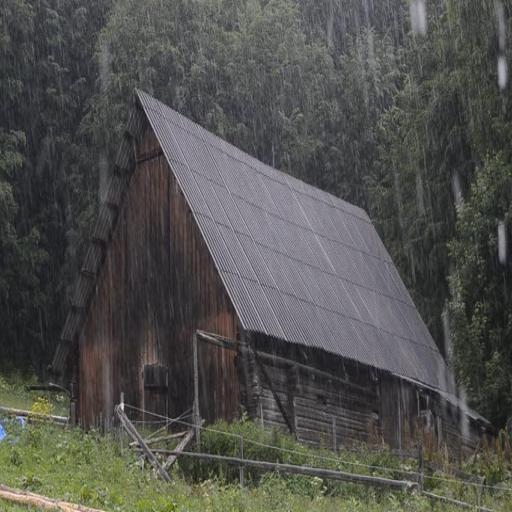}}
		\quad
		\subfigure[ PreNet ]{
			\label{prenet_real}
			\includegraphics[width=0.20\linewidth]{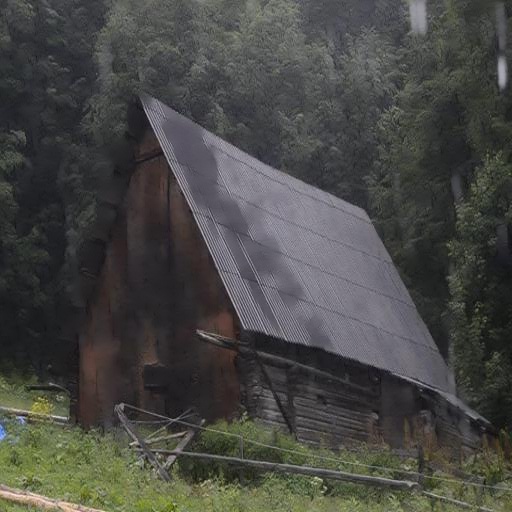}}
		\quad 
		\subfigure[ JORDER\_E ]{
			\label{jorder_real}
			\includegraphics[width=0.20\linewidth]{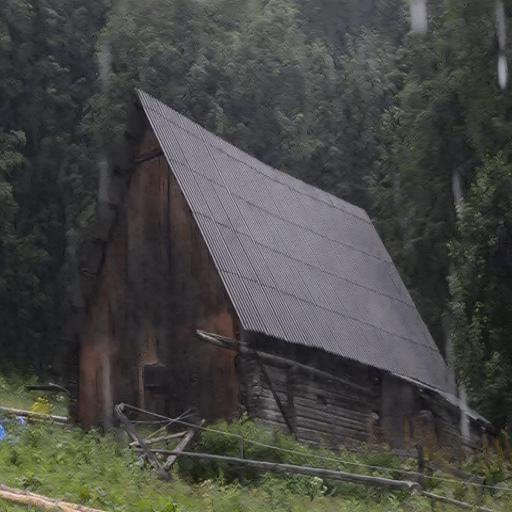}}
		\quad
		\subfigure[ Ours ]{
			\label{sift_real}
			\includegraphics[width=0.20\linewidth]{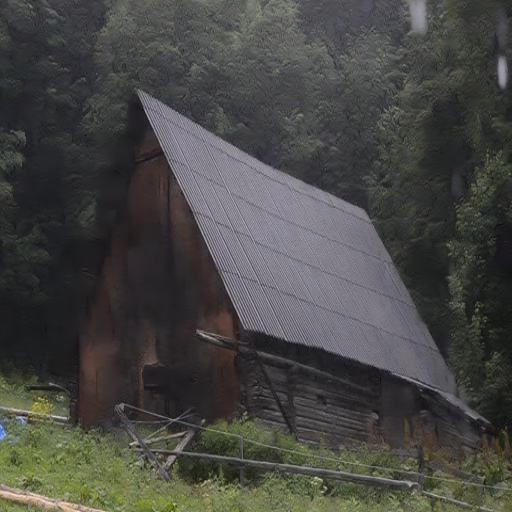}}
		
		\subfigure[ Input ]{
			\label{rain_real2}
			\includegraphics[width=0.20\linewidth]{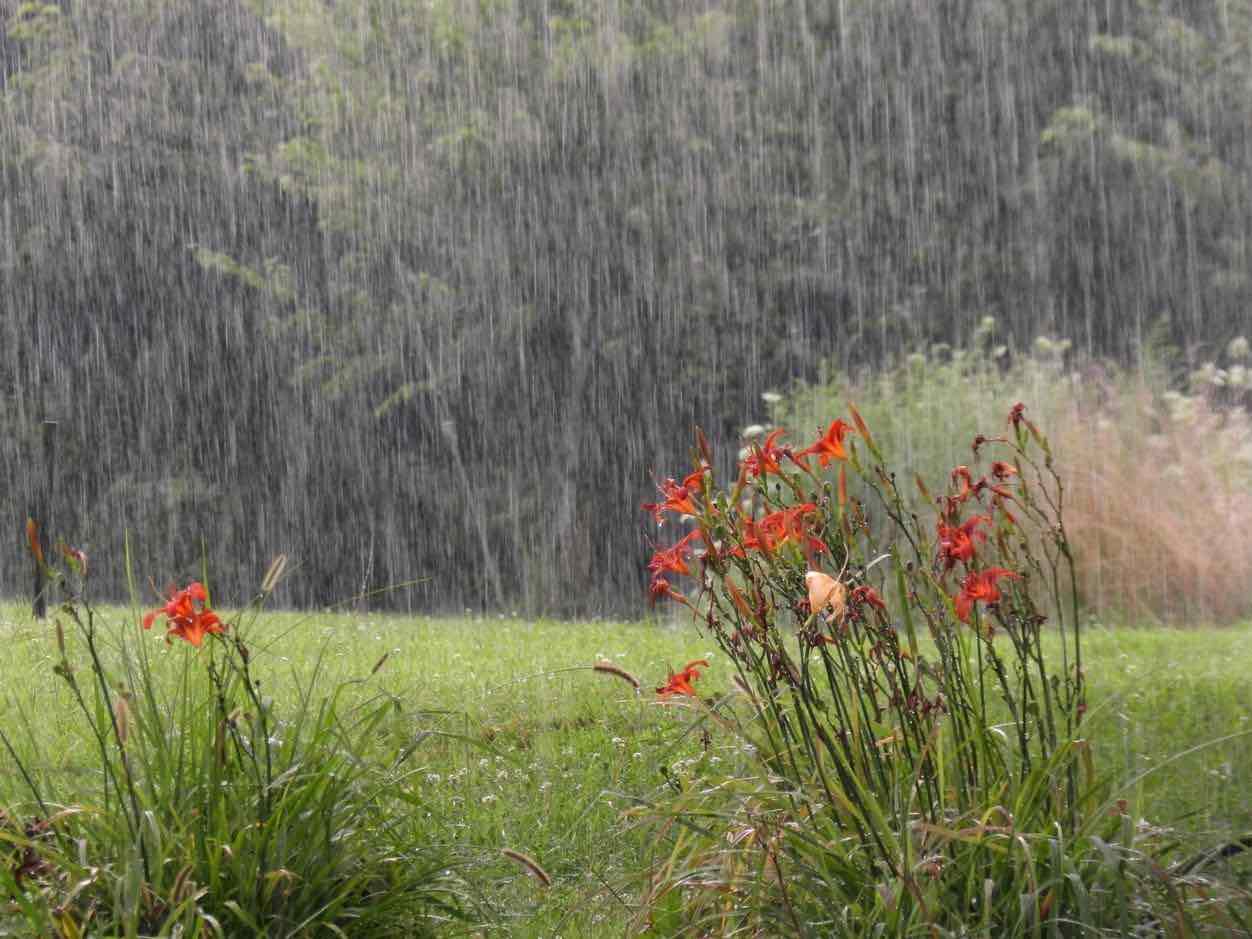}}
		\quad
		\subfigure[ PreNet ]{
			\label{prenet_real2}
			\includegraphics[width=0.20\linewidth]{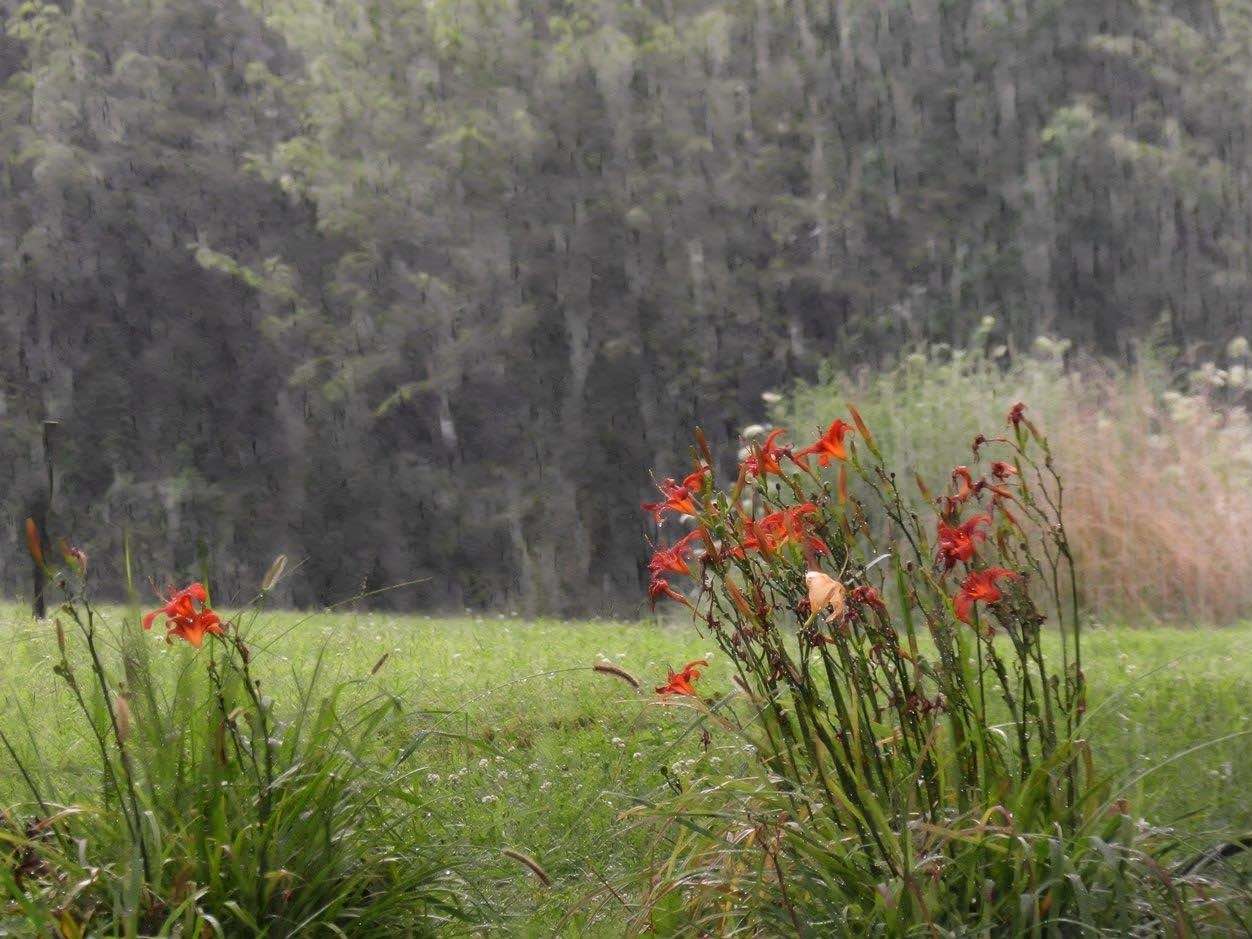}}
		\quad 
		\subfigure[ JORDER\_E ]{
			\label{jorder_real2}
			\includegraphics[width=0.20\linewidth]{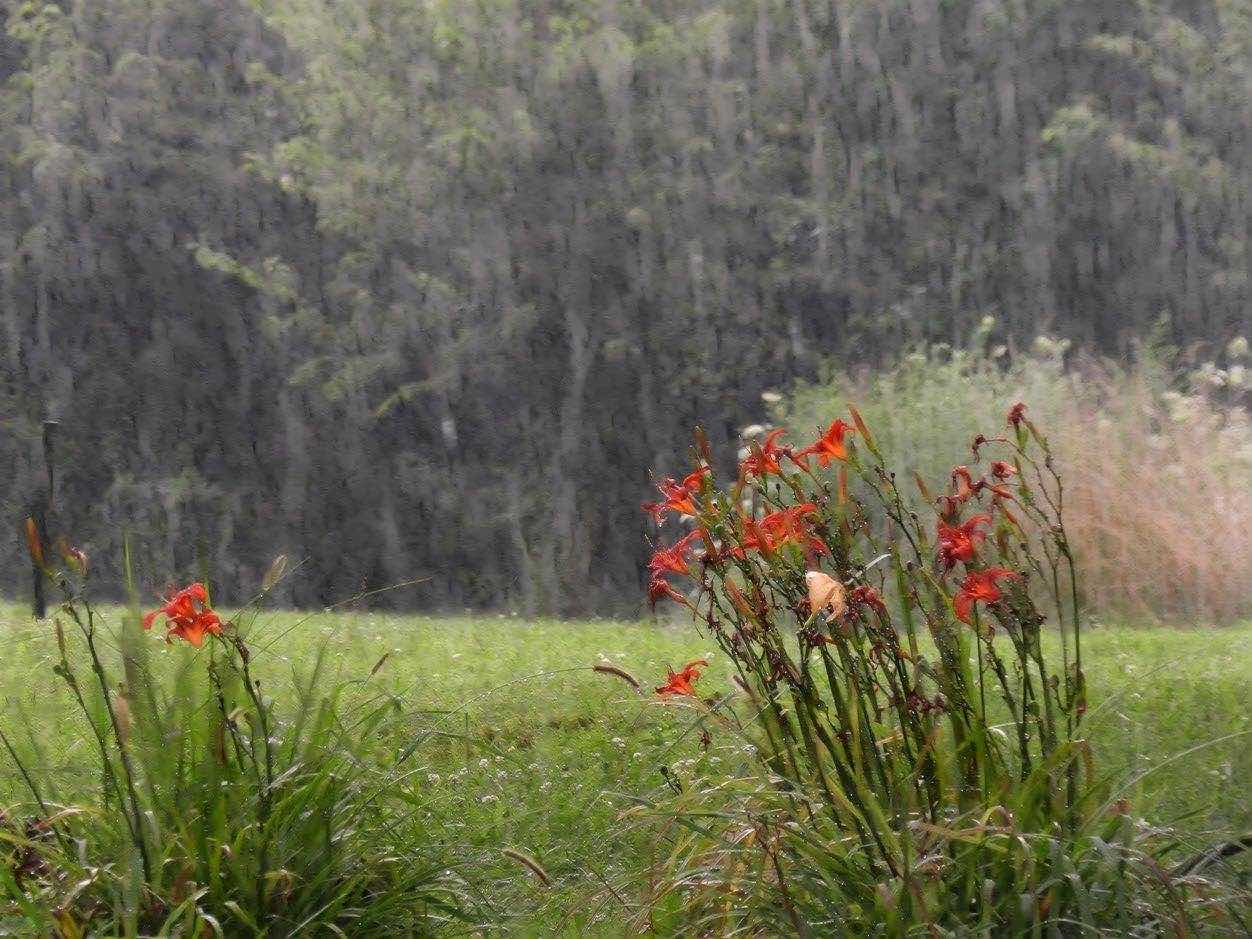}}
		\quad
		\subfigure[ Ours ]{
			\label{sift_real2}
			\includegraphics[width=0.20\linewidth]{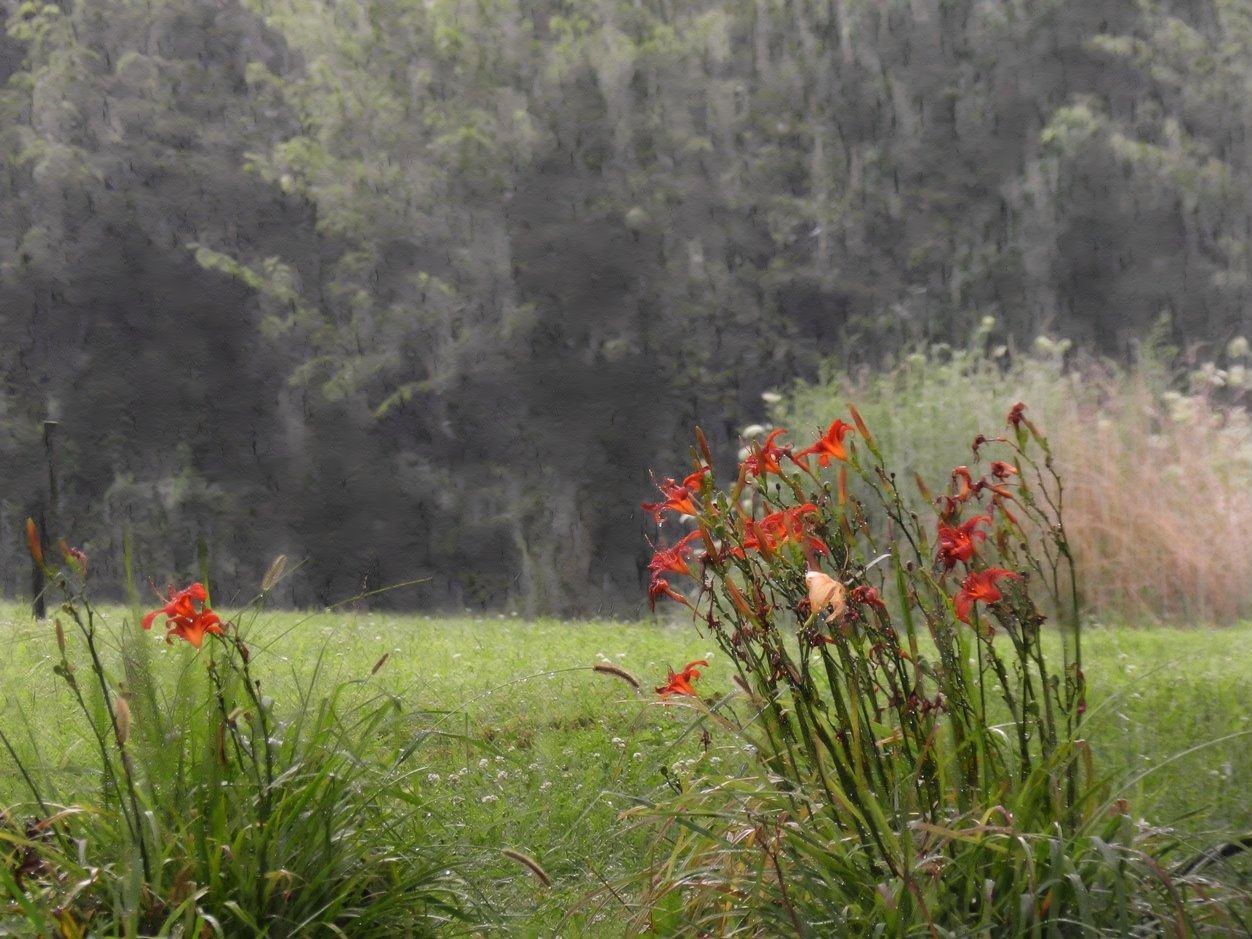}}
		
		\caption{Visual deraining results on real rainy images}
		\label{real}
	\end{figure*}
	
	\subsection{Evaluation on Real Rainy Scenes}
	We further investigate the performance of our method on real deraining cases. In Fig. \ref{real}, we show the subjective comparison results on two real rainy images with the PReNet and JORDER\_E. It can be found, for the first image, ours and JORDER\_E achieve better results than PReNet, which generates large oversmoothing regions on the roof. For the second image, the result by our scheme is more clear than the other two methods.

	\subsection{Ablation Study}
	The main modules of our scheme include multi-scale feature extraction, scale-space invariant attention network (SIAN), and LSTM based progressive refinement. In this subsection, we provide ablation analysis to show the roles of these modules to the final performance. We define the group of ablation study as:
	\begin{itemize}
		\item Baseline: only the first stage of Fig. \ref{fig:fig6} is used, and $\{\boldsymbol{M}_{(i)}\} = 0$, which means no attention mechanism is employed;
		\item Baseline+LSTM: all stages of Fig. \ref{fig:fig6} are used, and $\{\boldsymbol{M}_{(i)}\} = 0$; 
		\item Baseline+SIAN+LSTM: the complete form of Fig. \ref{fig:fig6}.
	\end{itemize}
	As shown in Table. \ref{ablation}, compared with Baseline, Baseline+LSTM works better. This demonstrate the strategy of progressive refinement is helpful. Compare with Baseline+LSTM, Baseline+SIAN+LSTM further improves the PSNR and SSIM performance, which demonstrates the proposed attention network really can improve the modeling ability of the network.
	
	
	\begin{table}[htbp]
		\caption{The ablation study about the role of modules to the final performance}
		\vspace{-0.3cm}
		\label{ablation}
		\begin{center}
			\setlength{\tabcolsep}{2mm}{
				\begin{tabular}{|c|c|c|c|c|}
					\hline
					\multirow{2}{*}{\textbf{Method}} & \multicolumn{2}{c|}{\textbf{Rain100L}}                                  & \multicolumn{2}{c|}{\textbf{Rain100H}}                                 \\ \cline{2-5} 
					& \multicolumn{1}{c|}{\textbf{PSNR}} & \multicolumn{1}{c|}{\textbf{SSIM}} & \multicolumn{1}{c|}{\textbf{PSNR}} & \multicolumn{1}{c|}{\textbf{SSIM}} \\ \hline
					Baseline                   & 37.92                              & 0.980                              & 28.55                              & 0.893                             \\ \hline
					Baseline+LSTM              & 38.68                              & 0.982                              & 30.20                              & 0.906                             \\ \hline
					Baseline+SIAN+LSTM         & \textbf{38.80}                     & \textbf{0.984}                     & \textbf{30.33}                     & \textbf{0.909}                    \\ \hline
				\end{tabular}
			}
		\end{center}
		\vspace{-0.6cm}
	\end{table}

	\section{Conclusion}
	In this work, we presented a novel single image deraining scheme based on scale-aware deep neural networks. To aggregate features from multiple scales into our rain steaks prediction, we developed a new scale-space invariant attention mechanism that learns a set of importance masks, one for each scale. Experimental results show that our proposed method achieves state-of-the-art performance with respect to both quantitative and qualitative evaluations.


	
\end{document}